\documentclass{article}

\usepackage{microtype}
\usepackage{graphicx}
\usepackage{subfigure}
\usepackage{booktabs} % for professional tables
\usepackage{listings}
\usepackage{tikz-cd}

\usepackage{hyperref}

\usepackage[accepted]{icml2025}
\usepackage{amsmath}
\usepackage{amssymb}
\usepackage{mathtools}
\usepackage{amsthm}
\usepackage[subtle]{savetrees}

\usepackage[capitalize,noabbrev]{cleveref}

\theoremstyle{plain}

\theoremstyle{definition}

\theoremstyle{remark}

\usepackage[textsize=tiny]{todonotes}

\icmltitlerunning{Using Enriched Category Theory to Construct the Nearest
Neighbour Classification Algorithm}

\begin{document}

\twocolumn[
\icmltitle{Using Enriched Category Theory to Construct the Nearest
Neighbour Classification Algorithm}



\icmlsetsymbol{equal}{*}

\begin{icmlauthorlist}
\icmlauthor{Matthew Pugh}{}
\icmlauthor{Jo Grundy}{}
\icmlauthor{Corina Cirstea}{}
\icmlauthor{Nick Harris}{}
\end{icmlauthorlist}

%\icmlaffiliation{yyy}{Department of XXX, University of YYY, Location, Country}
%\icmlaffiliation{comp}{Company Name, Location, Country}
%\icmlaffiliation{sch}{School of ZZZ, Institute of WWW, Location, Country}

\icmlcorrespondingauthor{Matthew Pugh}{mp8g16@soton.ac.uk}

\icmlkeywords{Machine Learning, polynomial, optimisation theory}

\vskip 0.3in
]





\begin{abstract}
This paper is the first to construct and motivate a Machine Learning algorithm solely with Enriched Category Theory, supplementing evidence that Category Theory can provide valuable insights into the construction and explainability of Machine Learning algorithms. It is shown that a series of reasonable assumptions about a dataset lead to the construction of the Nearest Neighbours Algorithm. This construction is produced as an extension of the original dataset using profunctors in the category of Lawvere metric spaces, leading to a definition of an Enriched Nearest Neighbours Algorithm, which, consequently, also produces an enriched form of the Voronoi diagram. Further investigation of the generalisations this construction induces demonstrates how the $k$ Nearest Neighbours Algorithm may also be produced. Moreover, how the new construction allows metrics on the classification labels to inform the outputs of the Enriched Nearest Neighbour Algorithm: Enabling soft classification boundaries and dependent classifications. This paper is intended to be accessible without any knowledge of Category Theory.
\end{abstract}
\section{Introduction}

Recent work has demonstrated that category theory may be useful in the analysis of machine learning algorithms\cite{shiebler_category_2021}. However, there are almost no examples of a category theoretic construction being used for both the model description and selection. Such a construction is necessary to allow future researchers to apply the tools of category theory in the analysis of machine learning algorithms.

This work has produced a construction of the Nearest Neighbours Algorithm (\textit{NNA}),which describes both its modelling behaviour and selection entirely within the language of Enriched Category Theory (\textit{ECT}). From this initial result, a series of observations about the structure of the algorithm can be made, which give an indication of how category theory may be used in the analysis of machine learning algorithms.

Firstly, the construction of the \textit{NNA} uses a category called Cost as its base of enrichment, which produces enriched categories that operate like metric spaces. This allows standard intuitions about the \textit{NNA} to inform the construction. However, no part of the construction depends on Cost being the base of enrichment. This makes the first outcome the production of an infinite family of algorithms for any choice of enrichment base, called \textit{V-NNA}.

Secondly it is shown that the k-Nearest Neighbours Algorithm (\textit{$k$-NNA}) can be constructed using the categorical description of the \textit{NNA}. It demonstrates that any \textit{$k$-NNA} may be presented as a \textit{NNA} when the data is projected into a metric space of tuples.

Thirdly, the construction of the \textit{NNA} allows not only the data points to inhabit a metric space, but also their class labels. This causes the construction of the \textit{NNA} to generalise the algorithms behaviour such that it may operate with soft or dependent classifications.

Finally, the category of cost enriched categories which the \textit{NNA} lives inside of is equipped with an internal language which may express concepts using distance, addition, truncated subtraction, infimum, and supremum. Because both model representation and selection occurs inside this category, the \textit{NNA} is only capable of understanding information expressible in these terms. The relationship between the internal language of the category and the behaviour of the \textit{NNA} has interesting implications for algorithm explainability.

\section{Background}

The construction of the Nearest Neighbours Algorithm demonstrated is the first examples of a machine learning algorithm constructed with Enriched Category Theory (\textit{ECT}). There is one other example of an entirely categorical construction of an ML algorithm, where previous work \citep{shieblerKanExtensionsData2022} shows that the single linkage clustering algorithm can be found as a Kan-extension of a dataset of points. 

There are examples of algorithms whose structures have been encoded in the language of category theory, such as Graph Neural Networks \citep{dudzik_graph_2022}. But they do not represent how the algorithm selects the best model.

It is beyond the scope of this paper to provide a complete introduction to Enriched Category Theory\footnote{A basic introduction can be found in "Seven Sketches in Compositionality" \citep{fongSevenSketchesCompositionality2018} while a more technical overview occurs in "Basic Concepts of Enriched Category Theory" \citep{Kelly2005}.}, but thankfully many of its complexities can be avoided by focusing on the specific case of Lawvere metric spaces.

\subsection{Nearest Neighbours Algorithm}
\label{section:NNA}
\begin{figure}
  \centering
  \includegraphics[width=0.5\linewidth]{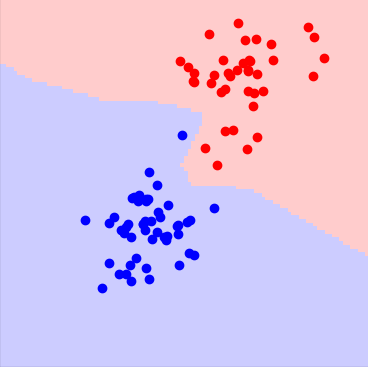}
  \caption{An example of the classification regions produced by the nearest neighbour algorithm from data points sampled from two Gaussian distributions, representing the distributions of the two classes.}
  \label{figure:NNA}
\end{figure}

The Nearest Neighbours Algorithm \citep{fix_discriminatory_1989} extends the classification of a dataset of points in a metric space to the entire metric space. Consider a dataset of $n$ pairs $(x_1, y_1), ... , (x_n, y_n)$. The targets of the dataset, $y_i$, are elements of a set of class labels $Y$. The features of the dataset, $x_i$, represent points in a metric space $X$. This allows the distance between any two points to be measured, following the traditional metric space axioms.
\begin{align*}
\bullet &\quad d(a,a) = 0 &&\\
\bullet &\quad a \neq b \Leftrightarrow d(a, b) > 0  &&\textit{Positivity} \\
\bullet &\quad d(a, b) = d(b, a) &&\textit{Symmetry} \\
\bullet &\quad d(a, b) + d(b, c) \geq d(a, c)  &&\textit{Triangle Inequality}
\end{align*}
To a point of the metric space not in the dataset, the Nearest Neighbours Algorithm assigns a class if a closest point in the dataset has that class. An example of the classification regions produced can be seen in Fig \ref{figure:NNA} which shows the \textit{NNA} classification of a two class dataset of points sampled from two Gaussian distributions.
The dataset may be represented with two functions. The indexes of the dataset can be expressed as the set of integers from 1 to n, $N = \{a \in \mathbf{Z}\ |\ 1 \leq a \leq n\}$. The features of the dataset can be encoded with the function $F : N \rightarrow X$ such that $Fi = x_i$. The targets of the dataset can be expressed similarly with a function $T : N \rightarrow Y$, such that $Ti= y_i$. Given a point $x\in X$ and a class $y \in Y$, the relation should return true if the closest data-point to $x$ has the class $y$. \footnote{$\inf$ in the following expression represents the infimum or least upper bound of a set of values. For finite cases it can be replaced with minimum.}
\begin{align}
\begin{split}
NNA(y, x) = \exists i \in N\ [\quad  &Ti = y\quad and\\
&\quad d(Fi, x) = \inf_{i' \in N}  d(Fi', x)\quad ]
\end{split}
\end{align}
Define the partition $NT(y) = \{ i \in N\ |\ Ti=y \}$. This allows the relation to be presented as: 
\begin{small}
\begin{equation}
NNA(y, x)\quad \Leftrightarrow \quad \inf_{i \in N}  d(Fi, x)\quad = \inf_{i \in NT(y)}  d(Fi, x)
\end{equation}
\end{small}
\subsection{Lawvere Metric Spaces}

Given an enriched category $C$, two objects $x \in C$ and $y \in C$ can be compared with the notation $C(x,y)$. This is referred to as the hom-object of $x$ and $y$. This hom-object exists in its own category called the base of enrichment. To make the comparisons meaningful, \textit{ECT}  requires that the base of enrichment have some way of combining hom-objects, called a monoidal product, and some juxtaposition of these two hom-objects to a third. An example of how this structure works can be seen in order relations. Consider a category called Fruits, which is a collection of fruits ordered by price. The hom-object $Fruits(Apple, Orange)$ would test to see if Apples were cheaper than Oranges. In this instance this comparison could also be written as $Apples \leq Oranges$. The outcome of this comparison is either true or false so the base of enrichment would be a category containing an object representing true and an object representing false. This base of enrichment can be called Bool for Boolean.

A sensible logical deduction to make with such a category would be to say that if we know fruit $A$ is cheaper than fruit $B$, and fruit $B$ is cheaper than fruit $C$, then $A$ must be cheaper than $C$. Notionally, this can be written as: \begin{equation} (A\leq B)\ and\ (B\leq C) \implies (A \leq C)\end{equation} This process of logical inference gives the general motivating structure of an enriched category. In this instance, each comparison of the ordered set returns a value in Bool. The monoidal product of Bool is the logical "and", allowing its objects to be combined. The hom-object of Bool is logical implication. \begin{equation} Bool(x,y) = (x \Rightarrow y) \end{equation}

%Bool also has arrows of implication from False to False, False to True, and True to True. But not from True to False, as True cannot logically imply False. 

Because Bool can be described with a hom-object which takes values in itself, it is described as self enriched. When Bool is used as the base of enrichment, the general structure of the enriched category becomes the structure of a pre-order relation. 

A Lawvere metric space is an enriched category whose base of enrichment is chosen so that the categories operate like metric spaces, allowing the enriched category to measure the distances between its objects. The base of enrichment for Lawvere metric spaces is called the Cost category. Because it represents measurements of distance, its objects are the non-negative real numbers extended with infinity\footnote{The objects of Cost being $\{x \in \mathbb{R}\ |\ x \geq 0 \} \cup \{\infty\}$. The monoidal product is addition, with addition by infinity defined as $x + \infty = \infty$}. Given a Cost enriched category $X$, and two objects $x$ and $y$ of $X$, the hom object $X(x,y)$ can be interpreted as the distance between $x$ and $y$. The monoidal product of Cost is addition. Cost can be interpreted as both a Bool enriched category, and a self enriched category. As a Bool enriched category, it is a preorder, with the ordering of its objects given by the standard order relation on the real numbers. \begin{equation} Cost(x,y) = (x \geq y)\end{equation} %and the arrows of the Cost category point from large numbers to smaller numbers. As in, there is an arrow from $a \in Cost$ to $b \in Cost$ if and only if $a\geq b$. This can also be interpreted as $Cost(a, b) = a \geq b$. 
Looking at the previous example, we can replace the $and$ operation of Bool with addition, and the implication with $\geq$ to recover the following expression for Cost categories. \begin{equation} X(x,y) + X(y, z) \geq X(x, z)\end{equation}
This requirement of Cost categories is the triangle inequality, stating that taking a detour to a third object cannot be quicker than traveling directly between two objects. By choosing the Cost category as the base of enrichment, \textit{ECT}  naturally recovers some, but not all of the the metric space axioms (As detailed in section \ref{section:NNA}). This makes Lawvere metric spaces pseudo-metric spaces. In Lawvere metric spaces, one retains the triangle inequality, and the requirement that the distance from an object to itself is zero ($d(a, a) = 0$), but the metric spaces are not required to be symmetric ($d(a,b) = d(b, a)$) and two different objects can be zero distance apart. This can be a controversial choice, but there are several arguments for this being a desirable outcome. For example, in many cases an intuitive notion of distance is not symmetric, e.g. its easier to go down stairs than up them. One might also say that distance is a measure of similarity not identity, and the idea of two different objects being zero distance apart is sensible when considering systems at a certain level of coarseness. In either case, if one wishes to operate with traditional metric spaces, they are all also Lawvere metric spaces, and the necessary axioms can be asserted as convenient.

%When considering what things can be represented as enriched categories, given a certain base of enrichment. An easy example is can always given as the base of enrichment itself. A base of enrichment is also self enriched. Returning to the example Bool, the categories enriched over Bool are pre-orders, they have some form of ordering relation. Considering that Bool contains the values of false and true, with arrows that encode implication, these values can also be considered as an ordering, such that $False \leq True$.

%\begin{align*}
%False \implies False\quad\quad\quad& False \leq False\\
%False \implies True \quad\quad\quad& False \leq True\\
%True \implies True \quad\quad\quad& True \leq True\\
%\end{align*}

%This is also the case for Cost. To recover the self enriched interpretation of cost

By sensibly considering how we wish to compare objects in our enriched categories, choosing objects and a monoidal product in the base of enrichment, we have recovered the structure of a metric space. Though the Lawvere metric space is one of the simpler examples of an enriched category, it starts to reveal the power of such a theory to construct complex structures for the representation of data.

\subsection{Functors and Profunctors}
\label{section:FunctorsAndProfunctors}

%An Enriched Category may be thought of as representing a particular datatype, with the structure of that datatype being represented by the hom-objects of the category. In order to interact with this information, there are many ways of comparing categories to each other. Between categories with the same base of enrichment, there are two constructions which are relevant for this work: Functors and Profunctors.
In set theory, a mapping from one set to another is called a function. In \textit{ECT} , there is a similar concept called a functor. Functors between enriched categories are structure preserving maps. In the case of Cost-enriched categories (Lawvere metric spaces), this reduces to the statement that functors are distance non-increasing functions. Given a functor $F:X\rightarrow Y$, from X to Y, this can be expressed as the statement that for any two objects $a,b \in X$. \begin{equation}X(a, b) \geq Y(Fa, Fb)\end{equation}
A set relation $R$ between two sets $X$ and $Y$ is often described as a subset of the Cartesian products of $X$ and $Y$, i.e. $R\subseteq X \times Y$. However, this relation can also be thought of as a function which returns true if the relation is true, and false if the relation is false: $R : Y \times X \rightarrow \{False, True\}$. In \textit{ECT} , this notion is extended to a functor from the tensor product of two categories to the base of enrichment. \begin{equation}R : Y^{op} \otimes X \rightarrow Cost\end{equation} Such a construction is called a profunctor. A profunctor $R : Y^{op} \otimes X \rightarrow Cost$, can be written as $R : X \nrightarrow Y$. The tensor product of two categories $Y^{op} \otimes X$ contains objects which are pairs of objects in $X$ and $Y$ similar to how the Cartesian product of sets contains pairs of elements of sets. The notation $Y^{op}$ is used to refer to a category with the objects of $Y$, but whose hom objects are reversed. \begin{equation}Y(a,b) = Y^{op}(b, a)\end{equation} 
With two set relations $R : X \nrightarrow Y$ and $S : Y \nrightarrow Z$, a composite relation can be produced of the form $ S \circ R : A \nrightarrow C$. The composition of two relations $R$ and $S$ is true for two inputs $x$ and $z$, if there exists an element $y$ in $Y$ such that $R(y, x)$ is true, and $S(z, y)$ is true. The logic of relation composition is described by the following equation.
\begin{equation}
 (S \circ R) (z, x) := \exists y\in Y[ R(y, x)\ and\ S(z, y) ]
\end{equation}
Similar to relations, profunctors can also be composed. Given Cost enriched profunctors $R : X \nrightarrow Y$ and $S : Y \nrightarrow Z$, the output of their composition bares a striking resemblance to the formula for relation composition.
\begin{equation}
 (S \circ R)(z, x) := \inf _{y\in Y}\ ( R(y, x) + S(z, y))
\end{equation}
In Boolean logic, the "and" operation outputs true only when both of its inputs are true, and false otherwise. In Cost enriched categories, a distance of zero can be interpreted as true, and a distance greater than zero is false. The sum of two values $a$ and $b$, where both are non-negative, can only be zero if both $a$ and $b$ are zero. In Cost category logic, $a+b$ is the logical "and" operation. Furthermore, the infimum operation is the Cost version of the existential quantifier. When $X$ is finite, The statement $\inf_{x \in X} Fx = 0$ means there exists a value $x$ such that $Fx$ is zero. In the infinite case, it suggests that there exists a value $Fx$ which is arbitrarily close to zero. Applying this logic to the definition of profunctors, it can be seen that profunctors produce truth values from pairs of objects. Such an interpretation can be represented by the functor $(0=x) : Cost \rightarrow Bool$. The following table highlights the comparable notions between standard Set theory and Lawvere metric spaces, though this comparison extends to all \textit{ECT}.
\vspace{-15pt}
\begin{center}
\resizebox{\columnwidth}{!}{
\begin{tabular}{ccc}
Logical Concept & Set Theory & Lawvere metric spaces\\
\hline
Truth Values & Bool & Cost\\
Conjunction & $\wedge$ & +\\
Mappings & Functions & Functors\\
Binary Predicate & Relations & Profunctors\\
Existential Quantifier & $\exists$ & $\inf$\\
Universal Quantifier & $\forall$ & $\sup$\\
\end{tabular}
}
\end{center}

It can be observed that functions are a special kind of relation, known as a functional relation. A function $F:N \rightarrow X$ is said to produce an element $Fi$ when given an element $i\in N$, but this behaviour can be represented directly as a relation $F_* : N \nrightarrow X$ which evaluates to a truth value under the condition $F_*(x,i) \Leftrightarrow (x = Fi)$. In fact, there is also a second relation of the opposite direction $F^*:X \nrightarrow N$ which represents the logical evaluation of the function $F^*(i, x) \Leftrightarrow (Fi = x)$. 

The interaction between functions and relations has a mirror in the interaction between functors and profunctors. A functor $F : N \rightarrow X$ canonically generates two profunctors. One of the same direction $F_* : N \nrightarrow X$ and one of the opposite direction $F^* : X \nrightarrow N$. They are defined with the aid of hom-objects, where $F_*(x, i) = X(x, Fi)$ and $F^*(i, x) = X(Fi, x)$. In the case of Lawvere metric spaces, the profunctors of $F$ evaluated on objects $x$ and $i$ can be read as: "The distance between $x$ and the image of $i$ under $F$".

\section{Constructing The Nearest Neighbours Algorithm}
\label{section:Construction}

Starting with a dataset of $n$ pairs $(x_1, y_1), ... , (x_n, y_n)$, the $x_i$ values are elements of a metric space $X$, and the $y_i$ values are class labels. Given a new point $x \in X$, what is the correct class label to associate with it?

An individual data point, $(x_i, y_i)$, has three components. An index value $i$, an associated point in the metric space $x_i$, and the classification label $y_i$. The $n$ index values, the points, and the class labels can be stored in the Cost-enriched category $N$, $X$, and $Y$. The information of the dataset can be represented by two functors. $F : N \rightarrow X$ maps the index values to their associated position in the metric space $x_i$. The functor $T : N \rightarrow Y$ maps data indexes to class labels.

In the case of the metric space $X$, it is clear that between any points $a, b \in X$, the hom object $X(a, b)$ should correspond directly with the distance metric on $X$. The objects of $N$, the indexes, possess no explicit relation to each other. This would suggest that the distances between indexes should be as un-constraining as possible. The lack of constraint would suggest that the Functors from $N$ to any other Cost category, should correspond directly with maps from the objects of $N$ to the other category. To achieve this, the category $N$ can be given the discrete metric.
\begin{equation}
N(i, j) =
\begin{cases} 
      0 & i = j \\
      \infty & i \neq j 
   \end{cases}
\end{equation}
As the objects of $N$ are maximally distance, any mapping of objects from $N$ satisfies the condition of being distance non increasing. The same logic can be applied to the objects of $Y$. Class labels should also have no meaningful relation to each other, so the discrete metric can be applied to $Y$ as well. The complete dataset can be represented by the following diagram.
%
% https://q.uiver.app/#q=WzAsMyxbMCwxLCJOIl0sWzEsMCwiWCJdLFsyLDEsIksiXSxbMCwyLCJUIiwyXSxbMCwxLCJGIl0sWzEsMiwiIiwyLHsic3R5bGUiOnsiYm9keSI6eyJuYW1lIjoiZG90dGVkIn19fV1d
\begin{equation}\begin{tikzcd}
	& X \\
	N && Y
	\arrow["T"', from=2-1, to=2-3]
	\arrow["F", from=2-1, to=1-2]
	\arrow[dotted, from=1-2, to=2-3]
\end{tikzcd}\end{equation}
To find the classes of all the points in $X$ would be to find a suitable candidate for the dotted arrow from $X$ to $Y$. Such an extension problem models the regression of $T$ on $X$. Finding a relation from dependent variables to independent variables. However, it is expected that two classification regions in $X$ may be touching, producing a boundary between classification regions which can have a trivially small distance. If the classes are assigned by functors, then the functors must be distance non-increasing. This would require that the classes in $Y$ have a distance of zero from each other, which would have an unfortunate consequence. The hom-objects are the only way to distinguish between objects of a category. Setting all of the distances between objects in $Y$ to zero would make all of the classes indistinguishable from each other in any categorical construction. It was correct to assign $Y$ the discrete metric, but not to expect the classifications to be represented by a functor. The classifications can in fact be represented by a profunctor $\textit{NNA}  : X \nrightarrow Y$.

The functors $F$ and $T$ both have two canonical profunctors associated with them. By selecting these profunctors appropriately, they may be composed to produce a profunctor from $X$ to $Y$.
%
% https://q.uiver.app/#q=WzAsMyxbMCwxLCJOIl0sWzEsMCwiWCJdLFsyLDEsIksiXSxbMCwxLCJGXyoiLDAseyJzdHlsZSI6eyJib2R5Ijp7Im5hbWUiOiJiYXJyZWQifX19XSxbMCwyLCJUIiwyLHsic3R5bGUiOnsiYm9keSI6eyJuYW1lIjoiYmFycmVkIn19fV0sWzEsMiwiIiwwLHsic3R5bGUiOnsiYm9keSI6eyJuYW1lIjoiZGFzaGVkIn19fV1d
\begin{equation}\begin{tikzcd}
	& X \\
	N && Y
	\arrow["{F^*}"', "\shortmid"{marking}, from=1-2, to=2-1]
	\arrow["T_*", "\shortmid"{marking}, from=2-1, to=2-3]
	\arrow["T_*\circ F^*", dashed, from=1-2, to=2-3]
\end{tikzcd}\end{equation}
The profunctor $F^* : X \nrightarrow N$ measures the distance between a point in $X$ and the image of a data point in $N$. The profunctor $T_* : N \nrightarrow Y$ does something similar, but because it is produced by a functor between discrete categories, its outputs are even easier to interpret. If a data index $i$ has a class $y$, i.e. $Ti = y$, then $T_*(y, i)$ will be $0$. However, if $i$ does not have class $y$ then $T_*(y, i)$ is infinity. Substituting these profunctors into the profunctor composition formula produces the following equation.
\begin{equation}
(T_*\circ F^*) (y, x) = \inf_{i \in N}\ (F^*(i, x) + T_*(y, i))
\end{equation}
If the class of $i$ selected by the infimum is not $y$, then $T(y, i)$ will be infinity, making the entire sum as large or larger than any other possible value. However, if the $i$ selected was of class $y$, then the formula returns $\inf_{i \in N} F^*(i, x)$. In other words, the composition $(T_*\circ F^*)(y, x)$ returns the distance from $x$ to the closest data point which is of class $y$. This could also be interpreted as evaluating the infimum of a partition of the indexes which have the class $y$ \footnote{Note that the following expression re-uses the notation $NT(y)$ introduced in section \ref{section:NNA} to represent the partition subset of $N$ with classes $y$, $NT(y) = \{i \in N\ |\ Ti = y\}$}.
\begin{equation}
(T_*\circ F^*)(y, x) = \inf_{i \in NT(y)}  d(Fi, x)
\end{equation}
In order to reproduce the \textit{NNA} the profunctor $T_*\circ F^*$ needs to be compared to a similar composition with a profunctor that has no knowledge of the classes, $\mathbf{1}_{NY} : N \nrightarrow Y$.

To model the notion that $\mathbf{1}_{NY}$ has no knowledge of the classes, it must respond true to any $i\in N$ and $y\in Y$, i.e. $\mathbf{1}_{NY}(y, i) = 0$ \footnote{This also makes $\mathbf{1}_{NY}$ the terminal profunctor of the category of profunctors between $N$ and $Y$, $Prof(N, Y)$}. Composing this profunctor with $F^*$ produces a composition with no knowledge of the classes.
\begin{align}
(\mathbf{1}_{NY} \circ F^*)(y, x) &= \inf_{i \in N}\ (F^*(i, x) + \mathbf{1}_{NY}(y, i)) \\
&= \inf_{i \in N} F^*(i, x)
\end{align}
Given a point $x \in X$ and class $y \in Y$, the profunctor $(\mathbf{1}_{NY} \circ F^*)(y, x)$ gives the distance to the closest point in the dataset (i.e. in the image of $F$). This composition has forgotten all class information. Comparing the outputs of both profunctors produces the \textit{NNA}. As their outputs are objects of the Cost category, the natural comparison is their hom-object in Cost.
\begin{align}
\textit{NNA} & : X \nrightarrow Y\\
\textit{NNA} &(y, x) := Cost((\mathbf{1}_{NY} \circ F^*)(y, x),\ (T_*\circ F^*)(y, x))
\end{align}
Because Cost can be viewed as a Bool enriched category, and therefore a preorder, this leads to the expression:
\begin{equation}
\textit{NNA} (y, x) = (\mathbf{1}_{NY} \circ F^*)(y, x)\ \geq\ (T_*\circ F^*)(y, x)
\end{equation}
A point $x$ is taken to have class $y$ when $\textit{NNA} (y, x)$ is true. Consider the situation that the closest data point $Fj$ to $x$ has class $y$, then $T(y, j) = 0$. The left hand side of the inequality finds the smallest distance from x to a data point with any class and the right hand side finds the smallest distance to a data point with class y. When the closest data point to $x$ has class $y$, the left hand side returns the same value as the right hand side and the inequality is true.
\begin{small}
\begin{align}
\textit{NNA} (y, x) & \Leftrightarrow Cost(\ (\mathbf{1}_{NY} \circ F^*)(y, x),\ (T_*\circ F^*)(y, x)\ )\\
    & \Leftrightarrow (\mathbf{1}_{NY} \circ F^*)(y, x) \quad \geq \quad (T_*\circ F^*)(y, x)\\
    & \Leftrightarrow \inf_{i \in N} F^*(i, x)\quad \geq \quad \inf_{i \in N}\ (F^*(i, x) + T(y, i))\\
    & \Leftrightarrow F^*(j, x)\quad \geq \quad F^*(j, x) + T(y, j)\\
    & \Leftrightarrow F^*(j, x)\ \geq\ F^*(j, x)\\
    & \Leftrightarrow True
\end{align}
\end{small}
Alternatively, in a situation where the nearest data point does not have class $y$, then $(T_*\circ F^*)(y, x) > (\mathbf{1}_{NY} \circ F^*)(y, x)$ and the output will be false. From this interpretation, it is clear that the \textit{NNA} profunctor produces the same classification as the Nearest Neighbours Algorithm. In its purely categorical form, the similarity between the profunctor construction and the relation introduced in Section \ref{section:NNA} is obscured, but it can be made clear through substitution.
\begin{small}
\begin{align}
\textit{NNA} (y, x) & \Leftrightarrow Cost(\ (\mathbf{1}_{NY} \circ F^*)(y, x)\ ,\ (T_*\circ F^*)(y, x)\  )\\
& \Leftrightarrow (\mathbf{1}_{NY} \circ F^*)(y, x)\quad \geq\quad (T_*\circ F^*)(y, x)\\
& \Leftrightarrow \inf_{i \in N} F^*(i, x)\quad \geq \quad \inf_{i\in N}\ (F^*(i, x) + T_*(y, i))\\
& \Leftrightarrow \inf_{i \in N} F^*(i, x)\quad \geq \quad \inf_{i\in NT(y)} F^*(i, x)\\
& \Leftrightarrow \inf_{i \in N} F^*(i, x)\quad = \quad \inf_{i\in NT(y)} F^*(i, x)\\
& \Leftrightarrow \inf_{i \in N} X(Fi, x)\quad = \quad \inf_{i\in NT(y)} X(Fi, x)\\
& \Leftrightarrow \inf_{i \in N} d(Fi, x)\quad = \quad \inf_{i\in NT(y)} d(Fi, x)
\end{align}
\end{small}
The last line is the same as the \textit{NNA}  relation shown in Section \ref{section:NNA}, demonstrating that this construction is the same as the standard Nearest Neighbours Algorithm.

\section{Generalising the Nearest Neighbours Algorithm}
\label{section:generalisations}

%With the basic construction of the \textit{NNA} presented in section \ref{section:Construction}, we now explore how this form can be generalised. Firstly, we will discuss the production of the k Nearest Neighbours Algorithm (\textit{$k$-NNA}). Secondly, we will discuss forming soft or dependent classification boundaries using non-discrete metrics for classification labels.

\subsection{K Nearest Neighbours}

The $k$-\textit{NNA} is an extension of the \textit{NNA} that bases its output on an aggregate classification formed from the classes of the $k$ nearest neighbours. To produce this generalisation, the construction must store the information associated with $k$ neighbouring data points, and a choice needs to be made concerning the method of aggregating the classes of these points. Given a Cost enriched category $C$, one can produce the $k$ times tensor product of $C$ with itself, using the tensor product introduced in section \ref{section:FunctorsAndProfunctors}. \begin{equation}C^k := C_1 \otimes ... \otimes C_k \quad\quad C = C_i\end{equation} The objects of $C^k$ are $k$-tuples of the objects of $C$. The tensor product operates on the hom objects of $C$ by applying the monoidal product of the base of enrichment. In the case of Cost, the monoidal product is addition. \begin{equation} C^k(\vec{x},\vec{y}) = C_1(\vec{x}_1, \vec{y}_1) \otimes ... \otimes C_k(\vec{x}_k, \vec{y}_k)\end{equation} The tensor product of functors maps tuples to tuples, with each component functor acting element-wise on the tuples. \begin{equation}F^k(\vec{x}) = (F(\vec{x}_1), ..., F(\vec{x}_k))\end{equation}
Because the tensor product of enriched categories acts on both categories and functors, one can produce the $k$ times tensor product of the diagram which defines the dataset.
%
% https://q.uiver.app/#q=WzAsMyxbMCwxLCJOXmsiXSxbMSwwLCJYXmsiXSxbMiwxLCJZXmsiXSxbMCwyLCJUXmsiLDJdLFswLDEsIkZeayJdXQ==
\begin{equation}\begin{tikzcd}
	& {X^k} \\
	{N^k} && {Y^k}
	\arrow["{F^k}", from=2-1, to=1-2]
	\arrow["{T^k}"', from=2-1, to=2-3]
\end{tikzcd}\end{equation}
This diagram can store the information required for the \textit{$k$-NNA} classification. However, it includes tuples that would usually be excluded, i.e., where a particular data point appears multiple times. Usually, the $k$ nearest neighbours are $k$ distinct neighbours. By including tuples which duplicate data points, the algorithm can base its aggregate classification on a tuple of duplicate points. To avoid this issue, we can introduce a subcategory of $N^k$, which only includes tuples with no duplicate points. This subcategory is equipped with a functor which is an inclusion on objects, mapping the subcategory into the original space, $I : \overline{N^k} \rightarrow N^k$. The introduction of $I$ allows the diagram to be restricted, by composing $I$ with $F$ and $T$, such that the algorithm only acknowledges tuples of distinct neighbours.  %then taking the induced profunctors, we can produce a similar profunctor diagram as used in section \ref{section:Construction}.
%
% https://q.uiver.app/#q=WzAsMyxbMCwxLCJcXG92ZXJsaW5le05ea30iXSxbMSwwLCJYXmsiXSxbMiwxLCJZXmsiXSxbMCwyLCIoVF5rSSlfKiIsMix7InN0eWxlIjp7ImJvZHkiOnsibmFtZSI6ImJhcnJlZCJ9fX1dLFsxLDAsIihGXmtJKV4qIiwyLHsic3R5bGUiOnsiYm9keSI6eyJuYW1lIjoiYmFycmVkIn19fV1d
\begin{equation}\begin{tikzcd}
	& {X^k} \\
	{\overline{N^k}} && {Y^k}
	\arrow["{(F^kI)^*}"', "\shortmid"{marking}, from=1-2, to=2-1]
	\arrow["{(T^kI)_*}"', "\shortmid"{marking}, from=2-1, to=2-3]
\end{tikzcd}\end{equation}
The second step in this procedure is to correct the outputs and inputs of the diagram by composing profunctors that convert $k$-tuples back into individual points of $X$ and $Y$. The $k$ times tensor product of $X$ with itself comes equipped with a profunctor functor that identifies the $k$-tuples formed from a single element, $\Delta : X \nrightarrow X^k$ The profunctor identifies such tuples by comparing a single element of $X$ with each element of the tuple using the distance metric of $X$. \begin{equation} \Delta(\vec{x}, x) = X(\vec{x}_1, x) \otimes ... \otimes X(\vec{x}_k, x) \end{equation} When every element of $\vec{x}$ is zero distance from $x$ then $\Delta(\vec{x}, x) = 0$. There are multiple viable options for the profunctor used to correct the outputs of the diagram. The aggregator profunctor takes the form $A : Y^k \nrightarrow Y$. Part of the variability in its definition is how $A$ handles ties, where the most common labels are equally prevalent amongst the $k$ neighbours. Whenever there is a clash, a greedy aggregation policy may assign both labels, a conservative policy would assign no labels, and a biased policy may always prefer one label over another. Ultimately, $A$ becomes a hyperparameter. The commonality between these schemes is that $A$ is invariant to permutations in the order of the elements, that it is $0$ when a tuple of neighbours does possess the given classification label, and it is $\infty$ otherwise. The following diagram may be produced with access to the profunctors $\Delta$ and $A$.
%
% https://q.uiver.app/#q=WzAsNSxbMCwxLCJcXG92ZXJsaW5le05ea30iXSxbMSwwLCJYXmsiXSxbMiwxLCJZXmsiXSxbMiwwLCJYIl0sWzMsMSwiWSJdLFswLDIsIihUXmtJKV8qIiwyLHsic3R5bGUiOnsiYm9keSI6eyJuYW1lIjoiYmFycmVkIn19fV0sWzEsMCwiKEZea0kpXioiLDIseyJzdHlsZSI6eyJib2R5Ijp7Im5hbWUiOiJiYXJyZWQifX19XSxbMywxLCJcXERlbHRhIiwyLHsic3R5bGUiOnsiYm9keSI6eyJuYW1lIjoiYmFycmVkIn19fV0sWzIsNCwiQSIsMix7InN0eWxlIjp7ImJvZHkiOnsibmFtZSI6ImJhcnJlZCJ9fX1dLFszLDQsIiIsMCx7InN0eWxlIjp7ImJvZHkiOnsibmFtZSI6ImRhc2hlZCJ9fX1dLFsxLDIsIiIsMSx7InN0eWxlIjp7ImJvZHkiOnsibmFtZSI6ImRhc2hlZCJ9fX1dXQ==
\begin{equation}\begin{tikzcd}
	& {X^k} & X \\
	{\overline{N^k}} && {Y^k} & Y
	\arrow["{(F^kI)^*}"', "\shortmid"{marking}, from=1-2, to=2-1]
	\arrow[dashed, from=1-2, to=2-3]
	\arrow["\Delta"', "\shortmid"{marking}, from=1-3, to=1-2]
	\arrow[dashed, from=1-3, to=2-4]
	\arrow["{(T^kI)_*}"', "\shortmid"{marking}, from=2-1, to=2-3]
	\arrow["A"', "\shortmid"{marking}, from=2-3, to=2-4]
\end{tikzcd}\end{equation}
In parallel with the strategy of section \ref{section:NNA} this allows the \textit{$k$-NNA} algorithm to be defined through profunctor composition.
\begin{small}
\begin{align}
\textit{$k$-NNA}& : X \nrightarrow Y\\
\textit{$k$-NNA}& := A \circ Cost((\mathbf{1}_{\overline{N^k}Y^k} \circ (FI)^*),\ ((TI)_*\circ (FI)^*))\circ \Delta
\end{align}
\end{small}
The definition of \textit{$k$-NNA} can also be presented more concisely with reference to the original \textit{NNA} algorithm, constructed over the functors $F^kI$ and $T^kI$. For clarity, we notate the original \textit{NNA} construction as $\textit{NNA}_{F,T}$.\begin{equation}\textit{$k$-NNA} = A \circ \textit{NNA}_{F^kI, T^kI} \circ \Delta\end{equation} Presenting the \textit{$k$-NNA} using the \textit{NNA} construction will also allow the arguments of section \ref{section:Construction} to be leveraged without having to repeat them. Each component profunctor must yield true for the composite \textit{$k$-NNA} to yield true. In the case of Cost-enriched profunctors, the output $0$ is interpretted interchangeably with the value true. To convince ourselves that the \textit{$k$-NNA} is behaving as expected, we may follow a point $x \in X$ around the diagram.

By construction, $\Delta(\vec{x}, x)$ is only true when $\vec{x}$ has as its elements objects which are zero distance from $x$. When $X$ is considered a traditional metric space, including the axiom of positivity, this forces $\vec{x}_i = x$. $\textit{NNA}_{F^kI, T^kI}(\vec{y}, \vec{x})$ is true when the closest point to $\vec{x}$ in the image of $F^kI$ has the classification $\vec{y}$. All points in the image of $F^kI$ are tuples of distinct neighbours. The distance metric of $X^k$ is the sum of distances in $X$. By minimising each dimension, we may construct an object $\vec{z} \in X^k$. We select $\vec{z}_1$ to be the closest point to $x$, and then as $\vec{z}_2$ must be a different point, we select it to be the second closest point in the dataset. Iterating leads to $\vec{z}_k$ being the $k$th closest point to $x$. This shows that if $\vec{z}$ minimises the distance to $\vec{x}$, it is a tuple of $k$ nearest neighbours to $x$. The classes of $\vec{y}$ must match component-wise with each class assigned to the components of $\vec{z}$. Given a $\vec{z}$ which minimises the distance to $\vec{x}$, then a permutation of $\vec{z}$ also minimises this distance, inducing a permutation in the classes $\vec{y}$. However, this does not affect the output because $A$ is selected to be invariant to permutations. Finally, given a tuple of classes $\vec{y}$, then $A(y, \vec{y})$ is true if $y$ corresponds with the aggregate class as selected by the chosen aggregation policy. The sequence of deductions shows that $\textit{$k$-NNA}(y, x)$ yields true only when $y$ is the aggregate of the classes of $k$ nearest neighbours to $x$.

In summary, \textit{$k$-NNA} can be produced using the \textit{NNA} algorithm by duplicating the original dataset with a $k$ fold tensor product. A choice of class label aggregation policy allows $k$-tuples of labels to be reduced to a single class label. %Before finishing this section, it should be noted that there are instances where a dataset may have multiple points which are equidistant from $x$. This would produce multiple tuples of neighbours, which minimise the distance without necessarily being a permutation of each other. In this case, the \textit{$k$-NNA} would assign multiple class labels to $x$. This can be seen as an extension of the case in the \textit{NNA}, where a point equidistant from two data points with different classes is assigned both class labels.

\subsection{Label Metrics for Soft Boundaries and Dependent Classifications}

%The categorical constructions of the \textit{NNA} and \textit{$k$-NNA} are novel to this paper, but the algorithms are already well established. While this is beneficial in justifying that \textit{ECT}  can produce practically applicable algorithms, it would be nice 
To demonstrate that \textit{ECT} can extend the utility of an existing algorithms it can be shown that the construction of the \textit{NNA} begins exhibiting novel and potentially valuable behaviours when its construction parameters are allowed to vary outside of what is usually possible.
%
%analysis can extend the utility of existing algorithms. Thankfully, the construction of the \textit{NNA} does not only replicate the behaviour of a traditional nearest neighbour algorithm. When the construction parameters are allowed to vary outside of what is usually possible, the \textit{NNA} begins exhibiting novel and potentially valuable behaviours.

The particular variation explored in this section occurs through the definition of the space of classification labels, $Y$. In the traditional implementation, the collection of classification labels is a set of values with no additional structure; as the algorithm is usually implemented directly from its description, it is unclear how one would incorporate additional structure into its decision-making. Section \ref{section:Construction} shows that when $Y$ is considered a discrete category, the \textit{ECT} construction of \textit{NNA} reproduces precisely the expected behaviour. However, $Y$ is not required to be a discrete category in general. By framing the construction within the category of Lawvere metric spaces, the enriched \textit{NNA} can accept $Y$ as any Lawvere metric space. Incorporating information about the similarity of classification labels as a metric allows the enriched classifier to produce softer classification boundaries. This could be formed from a semantic similarity metric for word-based labels or a more traditional metric for vector-based labels. Metrics associated with $Y$, the category of classification labels, will be referred to as label metrics.

\begin{figure}[ht]
  \centering
  \includegraphics[width=\linewidth]{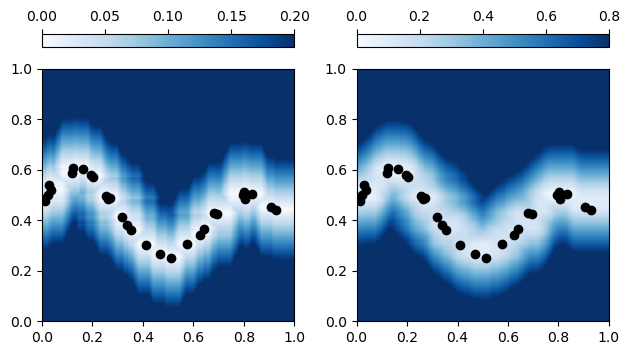}
  \caption{A plot of the values of $\textit{NNA}$ (left) and \textit{$4$-NNA} (right) for $x,y\in [0,1]$. $30$ points were uniformly sampled from the interval $[0,1]$ and transformed by the function $f(x) = 0.4+0.1\sin(10x)-0.7x^2 +0.7x^3$ then randomly scaled by $\pm5\%$ to produce the $Y$ values. The hom objects where taken to be $X(a,b) = Y(a,b) = |b-a|$. The aggregation policy chosen for \textit{$4$-NNA} is only zero when all tuple components agree on the class. Both colour scales clip outputs outside their stated range. }
  \label{figure:NNAreggresor}
\end{figure}

Label metrics are incorporated into the classification decision through $T_*$ as $T_*(y, i) = Y(y, Ti)$ When the label metric is discrete, $T_*$ serves only to identify when points are identical. However, for label metrics with bounded distances, the enriched \textit{NNA} may present a point as having a particular classification if that label is within a certain distance of the label of the nearest neighbour. Consider, for example, a function from the unit interval to the unit interval. A sample of points from this function can be presented with functors $F$ and $T$, allowing the application of the \textit{NNA}. Both $X$ and $Y$ values can be given the standard metric of the real numbers rather than forcing $Y$ to be discrete. The outputs of the \textit{NNA} and \textit{$4$-NNA} for every $X$ and $Y$ position are visualised in Fig \ref{figure:NNAreggresor}.

The images demonstrate the output values of the profunctors, showing how they evaluate the nearness to truth of each coordinate point. Operating in this fashion, the \textit{NNA} begins to act more as a regression model than as a classifier. 

The presentation of $Y$ as a Lawvere metric space also allows it to adopt an asymmetric metric. This induces what one might refer to as dependent classifications, where the assignment of one label may depend on the assignment of another. For example, consider the classification labels Dog, Cat, Bird, and Mammal. The label metric between Dog, Cat, and Bird can be taken to be discrete. As all Dogs and Cats are Mammals, the relationship can be represented by allowing the metric to become zero when evaluated in one direction. $Y(\textit{Mammal}, \textit{Dog}) = Y(\textit{Mammal}, \textit{Cat}) = 0$. As not all Mammals are Dogs or Cats, the converse relationship is assigned an infinite distance. $Y(\textit{Dog}, \textit{Mammal}) = Y(\textit{Cat}, \textit{Mammal}) = \infty$. As birds are not mammals and mammals are not birds, then these labels are an infinite distance from each other. $Y(\textit{Bird}, \textit{Mammal}) = Y(\textit{Mammal}, \textit{Bird}) = \infty$. Assume for some value $x$, that $\textit{NNA}(x, \textit{Dog}) = True$, then the two component profunctors of the \textit{NNA} are equal. $(\mathbf{1}_{NY} \circ F^*)(\textit{Dog}, x) = (T_*\circ F^*)(\textit{Dog}, x)$. The statement that $Y(\textit{Mammal}, \textit{Dog}) = 0$ allows it to be inserted into the definition of the right-hand side to demonstrate that $(\mathbf{1}_{NY} \circ F^*)(\textit{Dog}, x) = (T_*\circ F^*)(\textit{Mammal}, x)$.
\begin{small}
\begin{align}
(\mathbf{1}_{NY} \circ F^*)(\textit{Dog}, x) &= (T_*\circ F^*)(\textit{Dog}, x)\\
&= \inf_{i\in N}\ (F^*(i, x) + T_*(\textit{Dog}, i))\\
&= \inf_{i\in N}\ (F^*(i, x) + Y(\textit{Dog}, Ti))\\
\begin{split}
&=  \inf_{i\in N}\ (F^*(i, x) + Y(\textit{Mammal}, \textit{Dog})\\
&\quad\quad+Y(\textit{Dog}, Ti))
\end{split}\\
&\geq \inf_{i\in N}\ (F^*(i, x) + Y(\textit{Mammal}, Ti))\\
&\geq (T_*\circ F^*)(\textit{Mammal}, x)
\end{align}
\end{small}
The presence of the infimum in the definition of $(\mathbf{1}_{NY} \circ F^*)(\textit{Dog}, x)$ forces the inequality to be an equality. By the definition of the left-hand side, we know that $(\mathbf{1}_{NY} \circ F^*)(\textit{Dog}, x) = (\mathbf{1}_{NY} \circ F^*)(\textit{Mammal}, x)$ as $(\mathbf{1}_{NY} \circ F^*)$ is invariant to the first input. Together, these statements induce the following equality. \begin{equation}(\mathbf{1}_{NY} \circ F^*)(\textit{Mammal}, x) = (T_*\circ F^*)(\textit{Mammal}, x)\end{equation}This infers that $\textit{NNA}(\textit{Mammal}, x) = True$. Proving the aforementioned dependent classification. \begin{equation}\textit{NNA}(\textit{Dog}, x) \implies \textit{NNA}(\textit{Mammal}, x)\end{equation} The same logic also induces a dependent classification between Cat and Mammal. However, as the distance from Mammal to Dog or Cat is infinite, neither depends on the Mammal label. Similarly, as the labels Mammal and Bird are mutually infinite distance apart, neither depends on the other.

Exploring the enriched \textit{NNA} as $Y$ is allowed to take non-discrete and asymmetric metrics demonstrates that the resultant behaviours may apply to a broad range of use cases. When $Y$ is allowed to be non-discrete, the \textit{NNA} can make regression and classification decisions reminiscent of kernel density algorithms. When $Y$ is allowed to be asymmetric, it can encode dependent relationships reminiscent of simple ontologies. %In both cases, the enrichment of the \textit{NNA} has unlocked behaviours with machine learning applications inaccessible to the original algorithm.
\section{Conclusion}

The nascent field of Category Theory for Machine Learning has grown in recent years. As Category Theory is predominantly concerned with mathematical structure, there is hope that such techniques can improve our understanding of Machine Learning algorithms. Previous works have demonstrated that there is value in this avenue of research. However, there are insufficient examples. In response, this paper provides one of the first constructions of an existing algorithm in the language enriched category theory, where both its modelling behaviour and selection are presented within the language of category theory. The paper also demonstrates how such a construction might inform future analysis of machine learning algorithms with category theory.

The ability of category theory to generalise algorithm behaviour has been demonstrated in both the large and small scale. As no part of the construction relied directly on the use of $Cost$, this work has introduced an infinite family of algorithms for any choice of base of enrichment.
\begin{small}
\begin{align*}
\textit{V-NNA}&: X \nrightarrow Y\\
\textit{V-NNA}&(y, x) := V((\mathbf{1}_{NY} \circ F^*)(y, x),\ (T_*\circ F^*)(y, x)) 
\end{align*}
\end{small}
Because the algorithm was constructed in the category of Lawvere metric spaces, its classification labels can also be attributed with a metric, inducing the behaviour of soft and dependent classification labels.

The ability of category theory to make certain patterns and relationships more evident is also demonstrated. By representing the \textit{NNA} as a profunctor, it was possible to demonstrate how the behaviour of the \textit{$k$-NNA} is just another example of the \textit{NNA} when projected into a space of tuples. It was also possible to highlight the pattern of the extension triangle as a consequence of the description of the data as two functors. This can be seen as a specific example of the algebraic description of finding relation from independent to dependent variables, which will hopefully become more apparent as more examples of algorithms are found.

Finally, by representing the \textit{NNA} inside the category of Lawvere metric spaces, it becomes clearer how the environment of the algorithm affects its behaviour. As the internal language is expressible through distance, addition, truncated subtraction, infimum, and supremum, the \textit{NNA} is only capable of interacting with concepts expressed in these terms. The relationship bewteen an algorithms behaviours and the category it exists inside of has interesting implications for future work in machine learning explainability.

\bibliographystyle{icml2025}

\end{document}